\documentclass{article}
\usepackage{spconf,amsmath,graphicx}

\usepackage{multirow}
\usepackage{amsmath}
\usepackage{amsfonts}
\usepackage{amssymb}
\usepackage{pifont}

\title{CrowdFormer: Weakly-supervised Crowd counting with Improved Generalizability}
\name{Siddharth Singh Savner, Vivek Kanhangad}
\address{Department of Electrical Engineering, Indian Institute of Technology Indore, India}
\begin{document}
%
\noindent \copyright2022 IEEE. Personal use of this material is permitted. Permission from IEEE must be obtained for all other uses, in any current or future media, including reprinting/republishing this material for advertising or promotional purposes, creating new collective works, for resale or redistribution to servers or lists, or reuse of any copyrighted component of this work in other works.

\maketitle

\begin{abstract}
Convolutional neural networks (CNNs) have  dominated the field of computer vision for nearly a decade due to their strong
ability to learn local features. However, due to their limited receptive field, CNNs fail to model the global context. On the other hand, transformer, an attention-based architecture can model the global context easily. Despite this, there are limited studies that investigate the effectiveness of transformers in crowd counting. In addition, the majority of the existing crowd counting methods are based on the regression of density maps which requires point-level annotation of each person present in the scene. This annotation task is laborious and also error-prone. This has led to increased focus on weakly-supervised crowd counting methods which require only the count-level annotations. In this paper, we propose a weakly-supervised method for crowd counting using a pyramid vision transformer. We have  conducted extensive evaluations to validate the effectiveness of the proposed method. Our method is comparable to the state-of-the-art on the benchmark crowd datasets. More importantly, it shows remarkable generalizability. 
\end{abstract}
\section{Introduction}
\label{sec:intro}
Analysis of crowd scenes is an important task in video surveillance. Specifically, the estimation of the crowd motion, crowd behaviour, and population density are essential to enhancing public safety. A sub-task of crowd analysis is crowd counting which provides an estimate of the number of people in a crowd scene. In the last decade, CNN-based methods have proven to be effective for many computer vision tasks including crowd counting. Taking advantage of CNN's ability to learn powerful image representations, researchers have succeeded in generating comparatively better density maps than those generated using traditional hand-crafted features. 

Transformers, introduced by Vaswani et al. \cite{Vaswani} have achieved remarkable results for natural language processing (NLP) tasks. A transformer adopts a self-attention mechanism by differentially weighting the significance of each part of the input data. Motivated by the success of transformers in NLP, many researchers have explored transformers for computer vision tasks. Unlike the widely-used CNNs for computer vision tasks, transformers effectively capture long-range dependencies and have global receptive fields. Dosovitskiy et al. \cite{vit} introduced a vision transformer named ViT for image classification. The ViT first splits input images into patches and then projects these patches into 1-D sequences of linear embeddings. These embeddings are then fed into a transformer encoder. However, the output feature maps of ViT are of low resolution and single scale. More recently, hierarchical transformer architectures \cite{twins}, \cite{PVT}, and \cite{swin} have been introduced. These architectures provide multi-scale feature maps at the output, thereby overcoming the limitations of ViT. 

The main contributions of our work are as follows:
\begin{itemize}
\item We introduce a simple end-to-end crowd counting pipeline $-$ CrowdFormer that utilizes a pyramid structured vision transformer to extract multi-scale features with a global context.
\item We propose an effective feature aggregation module to combine features from different stages of the transformer and a simple regression head to estimate the crowd number. 
\item We conduct extensive experiments on four benchmark crowd counting datasets to demonstrate the effectiveness of the proposed method. Our method achieves state-of-the-art counting performance on two of the datasets. On the  remaining two datasets, our results are comparable to the state-of-the-art. Also, our approach has greater generalizability and achieves state-of-the-art performance in cross-dataset evaluations, outperforming even most of the fully-supervised methods.   
\end{itemize}
\begin{figure*}[!t]
    \centering
    \includegraphics[width=7.5in,height=2.9in]{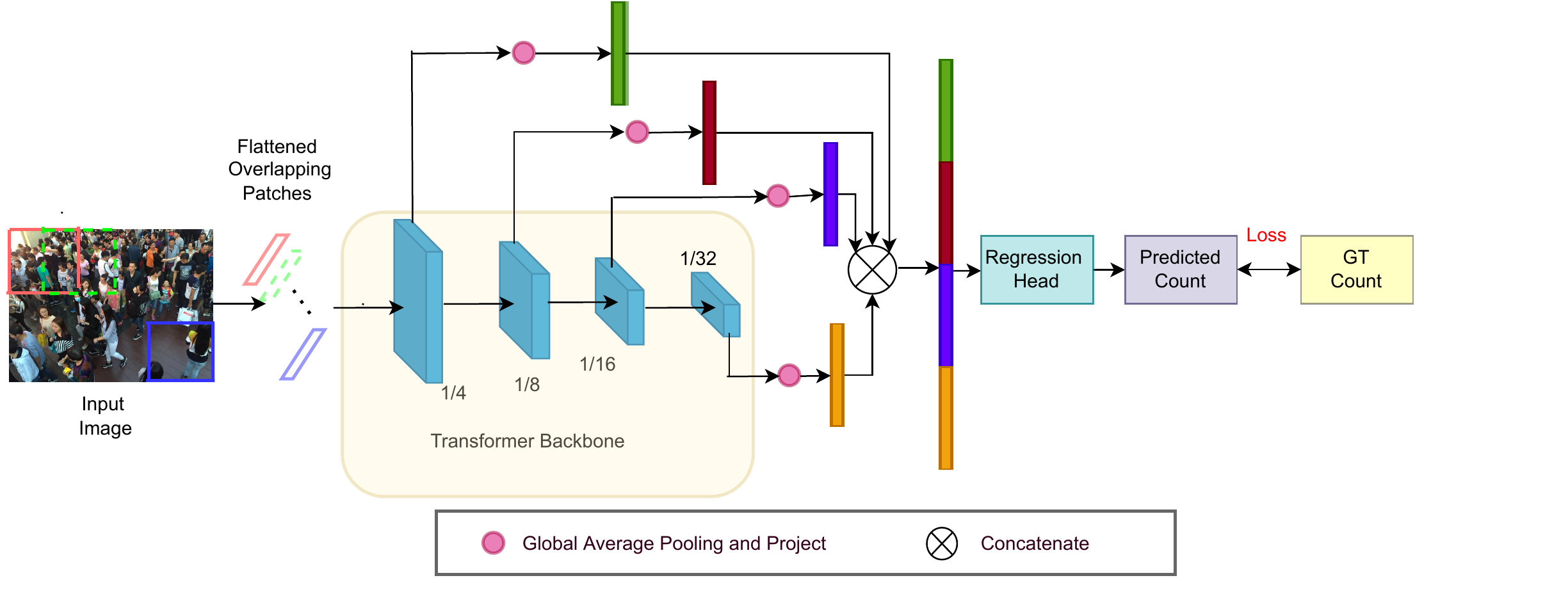}
    \caption{The pipeline of the proposed CrowdFormer}
    \label{fig:bd}
\end{figure*}
\section{Related Works}   
\label{sec:related_works}
In recent years, significant progress has been made with typical crowd counting methods that regress a density map. These density-map-regression-based methods employ a feature extraction module that extracts high-level features from the input images, followed by a regression head for pixel-level density map regression based on the extracted features. The drawback of these  methods is that they require point-level annotations to generate the ground truth density maps. Usually, the process of annotation involves placing a dot at the centre of the head of each person present in the image. This requires a lot of human effort and becomes challenging especially when the crowd density is high. In addition, human errors in labeling are inevitable. Furthermore, these annotations are not required at the time of evaluating the models. Therefore, it is desirable to develop weakly-supervised crowd counting methods, which require only the total number of people rather than the location of each person in the image.
 
In the literature, weakly-supervised crowd counting models have been developed without point-level annotations of the data or with limited amount of point-level annotations. The MATT \cite{MATT} is trained predominantly using count-level annotations, along with a small amount of point-level annotations. Similarly, Sam et al. \cite{Sam} optimized most of their model parameters with unlabeled data. The remaining few parameters were optimized using point-level annotations. The weakly-supervised solution in \cite{GaussianWeak} is based on the Gaussian process for estimating the crowd density. Yang et al. \cite{Yang}  proposed a soft-label sorting network that directly regresses the crowd count without any location supervision.
      
Transformer-based architectures have been explored in BCCT \cite{BCCT}, TransCrowd \cite{Transcrowd} and CCTrans \cite{CCtrans}. Among these, BCCT employs a ViT \cite{vit} based model for fully supervised crowd counting with various attention mechanisms. TransCrowd also utilizes a ViT based transformer encoder, followed by a regression head for weakly-supervised crowd counting. On the other hand, CCTrans \cite{CCtrans} uses Twins-SVT \cite{twins} backbone and a sophisticated decoder head. It can perform both fully as well as weakly-supervised crowd counting. 
\section{Proposed Method}
\label{sec:proposed_method}
The proposed pipeline for crowd counting is shown in Fig.\ref{fig:bd}. It employs a pre-trained transformer backbone to extract multi-scale features from the input image. The features from different stages are combined by a feature aggregation module and fed into a regression head which gives an estimate of the crowd count. 
\subsection{Transformer Backbone}
The proposed CrowdFormer uses a pyramid vision transformer PVTv2 \cite{PVTv2} as a feature extractor backbone. In particular, we use the `pvt\_v2\_b5' version of the PVTv2. It has four stages and each stage generates feature maps of a different scale. The architecture of each stage consists of an overlapping patch embedding layer and $L_i$ number of transformer encoder layers, i.e., $L_i$ encoder layers in $i^{th}$ stage. 

PVTv2 utilizes overlapping patch embeddings to tokenize images. While generating image patches, the adjacent windows are overlapped by half of its area. The overlapping patch embedding is implemented by applying a convolution with zero-padding and a suitable stride. Specifically, for an input of size $W\times H\times C$, a convolution layer with kernel size $2S-1$, zero-padding of $S-1$, stride $S$, and $C'$ number of kernels is used to generate an output of size $\frac{W}{S}\times\frac{H}{S}\times C'$. 
The stride in convolution for generating the patches is $S=4$ for the first stage and $S=2$ for the remaining stages. Thus, we obtain a set of feature maps from the $i^{th}$ stage, with dimensions scaled-down by $2^{(i+1)}$ as compared to the size of the input image.

At the beginning of stage $i$, the input is evenly divided into overlapping patches of equal size, then each patch is flattened and projected into $C_i$ dimensional embeddings. The embedding dimensions in PVTv2 are 64, 128, 320 and 512 for stage 1, 2, 3, and 4, respectively.  These patch embeddings are then passed through the transformer encoders. Each encoder consists of a self-attention mechanism and a feed-forward neural network. The position encoding is done in the feed-forward neural network. The transformer requires this positional information to determine the sequence order.

In CrowdFormer, the input image size is $384\times 384\times3$ pixels and the patch sizes are $7\times7\times3$ for the first stage and $3\times3\times C_i$, where $C_i$ is the embedding dimension of the $i^{th}$ stage. As mentioned previously, $C_2=64$, $C_3=128$, and $C_4=320$. Therefore, we get the output features (from individual stages) of dimension $96\times96\times64$, $48\times48\times128$,  $24\times24\times320$, and $12\times12\times512$.
\subsection{Feature Aggregation and Regression Head}
We have designed a novel feature aggregation module for effectively utilizing the outputs of the feature extractor backbone. We take the feature maps from each stage, perform global average pooling (GAP) to obtain 1-D sequences of dimensions 64, 128, 320, and 512. Each of these sequences is then projected into a 1-D sequence of length 6912. Further, these equal length sequences are concatenated to get a single sequence of length 27648. The resultant sequence is then fed into a regression head. We use a simple regression head consisting of only a single linear layer. It transforms the concatenated sequence into a scalar value representing the estimated crowd count. 
\subsection{Loss Function}
Since the crowd count varies significantly from one image to another, we have used smooth $L_1$ loss rather than $L_1$ loss, which is sensitive to outliers \cite{CCtrans}. The smooth $L_1$ loss is defined as follows: 
\begin{equation}
L_1\;\mbox{(smooth)} = 
     \begin{cases}
       \frac{0.5\times (x-y)^2}{\beta}, & \quad\text{if}\; |x-y|\le\beta\\     
       |x-y|-0.5\times\beta, & \quad\text{otherwise}\\
     \end{cases}
\end{equation}
The above loss function works as $L_1$-loss and $L_2$-loss when $|x-y|> \beta$ and $|x-y|\le\beta$, respectively. Here, $\beta$ is a hyper-parameter, $x$ and $y$ are the predicted and the ground truth crowd count, respectively. 

\section{Experiments}
\subsection{Training Setting and Hyper-parameters}
The proposed CrowdFormer uses a pre-trained transformer backbone for feature extraction. We have initialized the transformer encoder using the weights of the model pre-trained on the ImageNet-1K dataset \cite{imagenet1k} and followed the data preparation strategy adopted in \cite{Transcrowd}. During training, we use data augmentation techniques such as random horizontal flipping and grayscaling. Since image size varies within and across different datasets, all images are resized to $1152 \times 768$. Each resized image is partitioned into six non-overlapping images of size $384\times384$. 
	
We use AdamW \cite{AdamW} optimizer. The loss function parameter $\beta$ is set to 1 for SHA and QNRF, 7 for SHB, and 15 for UCF\_CC\_50  datasets. The learning rate and weight decay are set to 1e-5 and 1e-5, respectively. The batch size is set to 1. All our experiments are performed using the PyTorch framework on Google Colab with a single 16 GB P100 GPU.
\begin{table*}
    \centering
    \caption{Comparison with state-of-the-art weakly-supervised methods on SHA \cite{MCNN}, SHB \cite{MCNN}, UCF\_CC\_50 \cite{IdreesExtremelyDense}, and
QNRF \cite{UCF_QNRF} datasets. The best and second best results are shown in  \textbf{bold} and  \underline{underline}, respectively.
}
    \label{tab:the_table1}
    \begin{tabular}{|l|cc|cc|cc|cc|}
    \hline
    \multirow{2}{*}{Method}  & 
    \multicolumn{2}{c|}{UCF\_CC\_50}  &
    \multicolumn{2}{c|}{QNRF} & \multicolumn{2}{c|}{SHA} & \multicolumn{2}{c|}{SHB} \\
    \cline{2-9}
     &   MAE & MSE &  MAE & MSE &  MAE & MSE &  MAE & MSE\\
    \hline
    
    Yang et al. \cite{Yang} &-&-& - & - & 104.6 & 145.2 & 12.3 & 21.2\\
    MATT \cite{MATT} & 355.0&550.2& - & - & 80.1 & 129.4 & 11.7 & 17.5\\
    TransCrowd \cite{Transcrowd} &-&-&  97.2 & 168.5 & {66.1} & {105.1} & {9.3} & {16.1}\\      
    CCTrans \cite{CCtrans} &\underline{245.0}&\textbf{343.6}&  \textbf{92.1} &  \textbf{158.9}  &  \underline{64.4} &  \underline{95.4} & \textbf{ 7.0} &  \textbf{11.5}\\  
    \textbf{ours}  &\textbf{229.6}&\underline{360.3}& \underline{93.3} & \underline{160.9} &\textbf{62.1}&\textbf{94.8}  & \underline{8.5}  &\underline{13.6}\\
    \hline
    \end{tabular}
\end{table*}
\begin{table*}

    \centering
    \caption{Performance comparison of different methods in cross-dataset settings. The best and second best results are shown in  \textbf{bold} and  \underline{underline}, respectively. (\# - fully-supervised, * - weakly-supervised).
}
    \label{tab:the_table2}
    
    \begin{tabular}{|l|cc|cc|cc|cc|cc|cc|}
    \hline
    \multirow{2}{*}{Method} & 
    \multicolumn{2}{c|}{ SHB $\rightarrow$  SHA}  &
    \multicolumn{2}{c|}{{ SHA $\rightarrow$ SHB}} & 
    \multicolumn{2}{c|}{QNRF $\rightarrow$ SHA} & 
    \multicolumn{2}{c|}{QNRF $\rightarrow$  SHB} & 
    \multicolumn{2}{c|}{\small{SHA} $\rightarrow$ QNRF} & 
    \multicolumn{2}{c|}{\small{SHB} $\rightarrow$ QNRF} \\    
    \cline{2-13}
     &  MAE & MSE &  MAE & MSE &  MAE & MSE & MAE & MSE &  MAE & MSE &  MAE & MSE\\
    \hline
    MCNN \cite{MCNN}\#    &\underline{221.4} &  \underline{357.8} & 85.2 & 142.3 & - & - & -& -& -& -& -& -\\
    D-ConvNet \cite{DConvNet} \#
    & \textbf{140.4} & \textbf{226.1} & \underline{49.1} & \underline{99.2} & - & - & -& -& -& -& -& -\\    
    RRSP \cite{RRSP} \#  
    &- & - & \textbf{40.0} & \textbf{68.5} & - & - & -& -& -& -& -& -\\
   BL \cite{BL} \# 
    & - & - & -& - & \textbf{69.8} & \textbf{123.8} & \textbf{15.3} & \textbf{26.5} & -& -& -& -\\    
    \hline    

    TransCrowd \cite{Transcrowd}*  &\underline{141.3}&\underline{258.9}& \underline{18.9} & \underline{31.1} &\underline{78.7}&\underline{122.5}  & \underline{13.5}  &\underline{21.9}& -  & - &- & - \\
    
    \textbf{ours*}   &\textbf{121.6}&\textbf{208.8}& \textbf{16.0} & \textbf{26.0} & \textbf{70.8}&\textbf{114.9}  & \textbf{10.3}  &\textbf{16.0} &
    \textbf{147.6}  &\textbf{300.8} &
    \textbf{304.4}  &\textbf{586.1}\\
    \hline
    \end{tabular}
\end{table*}
\subsection{Datasets and Evaluation Metrics}
We have evaluated the proposed CrowdFormer on four benchmark datasets. The ShanghaiTech \cite{MCNN} dataset is a collection of 1,198 crowd images having a total of 330,165 annotations. This dataset is divided into two parts $-$ ShanghaiTech Part-A (SHA) containing 300 training and 182 testing images, and Part-B (SHB) containing 400 training and 316 testing images. 

The UCF-QNRF \cite{UCF_QNRF} (QNRF) dataset contains 1,535 images captured from various crowd scenes with approximately a million annotations. The count range varies from 49 to 12,865, with an average count of 815.4. The training and the testing set contain 1,201 and 334 images, respectively. 

The UCF\_CC\_50 \cite{IdreesExtremelyDense} is a very challenging dataset with severe perspective distortion. It is a collection of 50 gray-scale images collected from the Internet. It has a total of 63,974 annotations with an average people count of 1280. Since this dataset does not have a separate training and test set, we have performed 5-fold cross-validation by following the standard protocol discussed in \cite{IdreesExtremelyDense}.

As with previous works, we choose mean absolute error (MAE) and mean squared error (MSE) to evaluate the counting performance.
\begin{equation}
\mbox{MAE}=\frac{1}{N}\sum_{i=1}^{N}|P_i-G_i|
\end{equation}
\begin{equation}
\mbox{MSE}=\sqrt{\frac{1}{N}\sum_{i=1}^{N}|P_i-G_i|^2}
\end{equation}
where $N$ is the number of test images, $P_i$ and $G_i$ are the predicted and the ground truth count for the $i^{th}$ image, respectively.

\section{Results and Discussion}
\label{sec:results}
\subsection{Performance on Popular Benchmark Datasets}
To evaluate the performance of the proposed method, we conduct a set of experiments on four popular benchmark datasets. The results are presented in Table \ref{tab:the_table1}. Among the weakly-supervised methods, the proposed CrowdFormer clearly outperforms TransCrowd on all the datasets and achieves the state-of-the-art performance on SHA (in terms of both MAE and MSE) and  UCF\_CC\_50 (in terms of MAE). Specifically, CrowdFormer provides an improvement of 3.6\% in MAE and 0.63\% in MSE on SHA, 6.3\% in MAE on UCF\_CC\_50 compared with CCTrans \cite{CCtrans}. On the other datasets, our method achieves the second best results. 

The most plausible reason for superior performance of CrowdFormer compared with  state-of-the-art CNN-based crowd counting methods is its ability to capture the global context. As can be observed, CrowdFormer also outperforms another  transformer-based method TransCrowd \cite{Transcrowd} consistently. This is because TransCrowd uses only single-scale features, while our method learns multi-scale image features. Although CCTrans \cite{CCtrans} also employs a four-stage pyramid structured transformer to extract multi-scale features, it uses only features from the last three stages. In addition, it performs bilinear upsampling in its complex feature aggregation scheme to increase the resolution of the feature maps from the latter stages. This may have an impact on the quality of the feature maps. On the other hand, CrowdFormer uses features from all the four stages of the transformer and performs learnable projections during feature aggregation. Our method performs sequence to count mapping, which does not require summation of a 2-D density map to get the crowd count, as has been done in \cite{CCtrans}.
\subsection{Cross-dataset Performance}
We perform a set of cross-dataset evaluations on the four datasets to assess the generalizability of the proposed CrowdFormer. In these  evaluations, we train the model on a source dataset and test it on a target dataset without any fine-tuning. As can be observed in Table \ref{tab:the_table2}, CrowdFormer achieves state-of-the-art results for weakly-supervised crowd counting. Our method performs better than even fully supervised methods except for QNRF $\rightarrow$ SHA for which BL \cite{BL} achieves slightly lower MAE. Note that CCTrans \cite{CCtrans} has not reported its performance in any cross-dataset scenario. The results presented in Table \ref{tab:the_table2} indicate that the models trained on relatively more challenging dataset such as SHA or QNRF perform better on less challenging datasets such as SHB. As expected, the models trained on SHB do not perform very well on SHA and QNRF.

The CrowdFormer shows remarkable generalizability in cross-dataset evaluations due to its self-attention-like architecture that models short and long-range spatial dependencies \cite{Robustness}. A model having good generalizability is expected to perform well when deployed in the real-world and faced with unseen data with diverse characteristics arising from changes in crowd density, perspective, background, and scale.
\section{Conclusion}
 We have developed a pyramid vision transformer-based pipeline for crowd counting in a weakly-supervised setting. The transformer backbone extracts multi-scale features with global context and effectively captures the semantic crowd information. The feature aggregation module combines all the features followed by a simple regression head that regresses the crowd count. Extensive evaluations on four benchmark crowd datasets demonstrate the effectiveness of the proposed method in crowd counting and its generalizability.  



\bibliographystyle{IEEEtran}
\bibliography{main}

\begin{thebibliography}{10}
\providecommand{\url}[1]{#1}
\csname url@samestyle\endcsname
\providecommand{\newblock}{\relax}
\providecommand{\bibinfo}[2]{#2}
\providecommand{\BIBentrySTDinterwordspacing}{\spaceskip=0pt\relax}
\providecommand{\BIBentryALTinterwordstretchfactor}{4}
\providecommand{\BIBentryALTinterwordspacing}{\spaceskip=\fontdimen2\font plus
\BIBentryALTinterwordstretchfactor\fontdimen3\font minus
  \fontdimen4\font\relax}
\providecommand{\BIBforeignlanguage}[2]{{%
\expandafter\ifx\csname l@#1\endcsname\relax
\typeout{** WARNING: IEEEtran.bst: No hyphenation pattern has been}%
\typeout{** loaded for the language `#1'. Using the pattern for}%
\typeout{** the default language instead.}%
\else
\language=\csname l@#1\endcsname
\fi
#2}}
\providecommand{\BIBdecl}{\relax}
\BIBdecl

\bibitem{Vaswani}
A.~{Vaswani}, N.~{Shazeer}, N.~{Parmar}, J.~{Uszkoreit}, L.~{Jones}, A.~N.
  {Gomez}, L.~{Kaiser}, and I.~{Polosukhin}, ``Attention is all you need,'' in
  \emph{Advances in Neural Information Processing Systems}, 2017, pp.
  5998--6008.

\bibitem{vit}
A.~{Dosovitskiy}, L.~{Beyer}, A.~{Kolesnikov}, D.~{Weissenborn}, X.~{Zhai},
  T.~{Unterthiner}, M.~{Dehghani}, M.~{Minderer}, G.~{Heigold}, S.~{Gelly},
  J.~{Uszkoreit}, and N.~{Houlsby}, ``An image is worth 16x16 words:
  Transformers for image recognition at scale,'' in \emph{International
  Conference on Learning Representations}, 2021.

\bibitem{twins}
X.~Chu, Z.~Tian, Y.~Wang, B.~Zhang, H.~Ren, X.~Wei, H.~Xia, and C.~Shen,
  ``Twins: Revisiting the design of spatial attention in vision transformers,''
  in \emph{NeurIPS}, 2021.

\bibitem{PVT}
W.~Wang, E.~Xie, X.~Li, D.-P. Fan, K.~Song, D.~Liang, T.~Lu, P.~Luo, and
  L.~Shao, ``Pyramid vision transformer: A versatile backbone for dense
  prediction without convolutions,'' in \emph{IEEE/CVF International Conference
  on Computer Vision (CVPR)}, 2021, pp. 568--578.

\bibitem{swin}
Z.~Liu, Y.~Lin, Y.~Cao, H.~Hu, Y.~Wei, Z.~Zhang, S.~Lin, and B.~Guo, ``Swin
  transformer: Hierarchical vision transformer using shifted windows,''
  \emph{International Conference on Computer Vision (ICCV)}, 2021.

\bibitem{MATT}
Y.~Lei, Y.~Liu, P.~Zhang, and L.~Liu, ``Towards using count-level weak
  supervision for crowd counting,'' \emph{Pattern Recognition}, vol. 109, p.
  107616, 2020.

\bibitem{Sam}
D.~Sam, N.~Sajjan, H.~Maurya, and R.~Babu, ``Almost unsupervised learning for
  dense crowd counting,'' \emph{Proceedings of the AAAI Conference on
  Artificial Intelligence}, vol.~33, pp. 8868--8875, 2019.

\bibitem{GaussianWeak}
M.~Borstel, M.~Kandemir, P.~Schmidt, M.~Rao, K.~Rajamani, and F.~Hamprecht,
  ``Gaussian process density counting from weak supervision,'' in \emph{2016
  European Conference on Computer Vision (ECCV)}, 2016, pp. 365--380.

\bibitem{Yang}
Y.~Yang, G.~Li, Z.~Wu, L.~Su, and N.~Sebe, ``Weakly-supervised crowd counting
  learns from sorting rather than locations,'' in \emph{2020 European
  Conference on Computer Vision (ECCV)}, 11 2020, pp. 1--17.

\bibitem{BCCT}
G.~Sun, Y.~Liu, T.~Probst, D.~Pani~Paudel, N.~Popovic, and L.~Van~Gool,
  ``Boosting crowd counting with transformers,'' \emph{arXiv e-prints}, pp.
  arXiv--2105, 2021.

\bibitem{Transcrowd}
D.~{Liang}, X.~{Chen}, W.~{Xu}, Y.~{Zhou}, and X.~{Bai}, ``Trans{C}rowd:
  Weakly-supervised crowd counting with transformer,'' \emph{arXiv preprint
  arXiv:2104.09116}, 2021.

\bibitem{CCtrans}
Y.~{Tian}, X.~{Chu}, and H.~{Wang}, ``{CCT}rans: Simplifying and improving
  crowd counting with transformer,'' \emph{arXiv preprint arXiv:2109.14483},
  2021.

\bibitem{PVTv2}
W.~Wang, E.~Xie, X.~Li, D.-P. Fan, K.~Song, D.~Liang, T.~Lu, P.~Luo, and
  L.~Shao, ``{PVT}v2: Improved baselines with pyramid vision transformer,''
  \emph{Computational Visual Media}, vol.~8, no.~3, pp. 1--10, 2022.

\bibitem{imagenet1k}
O.~Russakovsky, J.~Deng, H.~Su, J.~Krause, S.~Satheesh, S.~Ma, Z.~Huang,
  A.~Karpathy, A.~Khosla, M.~Bernstein \emph{et~al.}, ``Imagenet large scale
  visual recognition challenge,'' \emph{International journal of computer
  vision}, vol. 115, no.~3, pp. 211--252, 2015.

\bibitem{AdamW}
I.~Loshchilov and F.~Hutter, ``Decoupled weight decay regularization,'' in
  \emph{International Conference on Learning Representations, {ICLR}}, 2019.

\bibitem{MCNN}
Y.~{Zhang}, D.~{Zhou}, S.~{Chen}, S.~{Gao}, and Y.~{Ma}, ``Single-image crowd
  counting via multi-column convolutional neural network,'' in \emph{IEEE
  Conference on Computer Vision and Pattern Recognition (CVPR)}, 2016, pp.
  589--597.

\bibitem{IdreesExtremelyDense}
H.~{Idrees}, I.~{Saleemi}, C.~{Seibert}, and M.~{Shah}, ``Multi-source
  multi-scale counting in extremely dense crowd images,'' in \emph{IEEE
  Conference on Computer Vision and Pattern Recognition (CVPR)}, 2013, pp.
  2547--2554.

\bibitem{UCF_QNRF}
H.~Idrees, M.~Tayyab, K.~Athrey, D.~Zhang, S.~Al-Maadeed, N.~Rajpoot, and
  M.~Shah, ``Composition loss for counting, density map estimation and
  localization in dense crowds,'' in \emph{European Conference on Computer
  Vision (ECCV)}, 2018, pp. 532--546.

\bibitem{DConvNet}
Z.~Shi, L.~Zhang, Y.~Liu, X.~Cao, Y.~Ye, M.-M. Cheng, and G.~Zheng, ``Crowd
  counting with deep negative correlation learning,'' in \emph{2018 IEEE/CVF
  Conference on Computer Vision and Pattern Recognition (CVPR)}, 2018, pp.
  5382--5390.

\bibitem{RRSP}
J.~Wan, W.~Luo, B.~Wu, A.~B. Chan, and W.~Liu, ``Residual regression with
  semantic prior for crowd counting,'' in \emph{2019 IEEE/CVF Conference on
  Computer Vision and Pattern Recognition (CVPR)}, 2019, pp. 4031--4040.

\bibitem{BL}
Z.~Ma, X.~Wei, X.~Hong, and Y.~Gong, ``Bayesian loss for crowd count estimation
  with point supervision,'' in \emph{2019 IEEE/CVF International Conference on
  Computer Vision (ICCV)}, 2019, pp. 6141--6150.

\bibitem{Robustness}
\BIBentryALTinterwordspacing
Y.~Bai, J.~Mei, A.~L. Yuille, and C.~Xie, ``Are transformers more robust than
  {CNN}s?'' \emph{CoRR}, vol. abs/2111.05464, 2021. [Online]. Available:
  \url{https://arxiv.org/abs/2111.05464}
\BIBentrySTDinterwordspacing

\end{thebibliography}

\end{document}